\documentclass[journal,twoside,web]{IEEEtran}
\usepackage{graphicx}
\usepackage{cite}
\usepackage{amsmath,amssymb,amsfonts}
\usepackage{algorithm}
\usepackage{algorithmic}
\usepackage{textcomp}
\usepackage{booktabs}
\usepackage{color,soul}
\usepackage{xspace}
\usepackage{xcolor}
\usepackage[colorlinks]{hyperref}
\usepackage{amsmath}
\usepackage{footnote}
\usepackage{makecell}
\usepackage{multirow}
\usepackage[export]{adjustbox}

\usepackage{array}
\usepackage[nolist,nohyperlinks]{acronym}
\usepackage{textcomp}

\makesavenoteenv{tabular}

\usepackage[acronym]{glossaries}

\acrodef{CT}{Computed Tomography}
\acrodef{CNN}{Convolutional Neural Network}
\acrodef{ASPP}{Atrous Spatial Pyramid Pooling}
\acrodef{mIoU}{mean Intersection over Union}
\acrodef{DSC}{Dice Coefficient}
\acrodef{FCN}{Fully Convolutional Network}
\acrodef{SOTA}{State Of The Art}
\acrodef{DSDF}{Dual-Scale Dense Fusion}
\acrodef{MSRF}{Multi-Scale Residual Fusion}
\acrodef{DSB}{Data Science Bowl}
\acrodef{MRI}{Magnetic Resonance Imaging}
\acrodef{PPM}{Pyramid Pooling Module}
\acrodef{FPS}{Frames Per Second}

\begin{document}
\title{SSTFB: Leveraging self-supervised pretext learning and temporal self-attention with feature branching for real-time video polyp segmentation}
\author{Ziang Xu, Jens Rittscher, and
       Sharib Ali
\thanks{Z. Xu and J. Rittscher are with the Institute of Biomedical Engineering, Department of Engineering Science, University of Oxford, OX3 7DQ, Oxford, United Kingdom (emails: \{ziang.xu, jens.rittscher\} @eng.ox.ac.uk)}
\thanks{S. Ali is with the School of Computing, Faculty of Engineering and Physical Sciences, University of Leeds, LS2 9JT, Leeds, United Kingdom (corresponding email: s.s.ali@leeds.ac.uk)}
}


\maketitle
\begin{abstract}
Polyps are early cancer indicators, so assessing occurrences of polyps and their removal is critical. They are observed through a colonoscopy screening procedure that generates a stream of video frames. Segmenting polyps in their natural video screening procedure has several challenges, such as the co-existence of imaging artefacts, motion blur, and floating debris. Most existing polyp segmentation algorithms are developed on curated still image datasets that do not represent real-world colonoscopy. Their performance often degrades on video data. We propose a video polyp segmentation method that performs self-supervised learning as an auxiliary task and a spatial-temporal self-attention mechanism for improved representation learning. Our end-to-end configuration and joint optimisation of losses enable the network to learn more discriminative contextual features in videos. Our experimental results demonstrate an improvement with respect to several state-of-the-art (SOTA) methods. Our ablation study also confirms that the choice of the proposed joint end-to-end training improves network accuracy by over 3\% and nearly 10\% on both the Dice similarity coefficient and intersection-over-union compared to the recently proposed method PNS+ and Polyp-PVT, respectively. Results on previously unseen video data indicate that the proposed method generalises. 

\end{abstract}


\IEEEpeerreviewmaketitle

\section{Introduction}
\label{sec:introduction}
\IEEEPARstart{C}olorectal cancer (CRC) is the third most common cancer worldwide~\cite{silva2014toward}. Early screening and removal of precancerous lesions (colorectal adenomas) can be effective in preventing the development of colorectal cancer and improving survival rates for CRC patients. Evaluation of polyps is an important step in cancer prevention. Several studies~\cite{haggar2009colorectal} provide evidence that survival rates for patients with stage IV and V CRC are reduced by 35\% compared to stage I. As the gold standard for CRC screening, colonoscopy has been widely used in clinical practice. During a colonoscopy, an endoscopist confirms the location of rectal polyps and assesses their visual appearance. A comprehensive visual assessment of polyps could eventually reduce the need for unnecessary biopsies, hence reducing the workload for pathologists and impacting the wider patient pathway. However, endoscopy is highly operator-dependent, and many polyps are missed during screening. Computer-assisted approaches are needed to improve diagnostic quality. Here, the accurate segmentation of polyps in real-time is a critical step. 

Polyps are visually very similar to their surrounding mucosa, and the low contrast affects the effectiveness of most segmentation algorithms. Constant camera motion and artefacts make this task even more challenging. Effective segmentation methods must compensate for motion blur, changes in illumination, occlusion, and specular reflections. While most methods are built upon image-based segmentation techniques~\cite{fang2019selective, fan2020pranet,dong2021polyp}, in the real world, the nature of the colonoscopic procedure is video. Hence, it is crucial to develop real-time video polyp segmentation that leverages the samples from previous frames, tackling false positives. Effectively, this can also improve the robustness of methods when the polyps are occluded or covered with artefacts such as stool. To this end, there are only a few methods that have developed real-time video segmentation. However, this method suffers from generalisability and limits its performance as they are solely based on fully supervised learning. Self-supervised learning (SSL) methods~\cite{jaiswal2020survey} have proven to be efficient ways of learning representations by minimising the dissimilarity between the samples. However, the question we have asked is - could we leverage the representation learning as an auxiliary task and incorporate the learnt representations to strengthen the supervised approach? Our work builds upon~\cite{ji2022video} but incorporates several novel ideas that include - the exploitation of the semantic meaning of frames using an SSL loss within the learning framework, and additionally, we exploit the sub-branching of the feature representations. 
\section{Related work}

\subsection{Medical image segmentation}
The visual similarity of polyps and the colonic mucosa severely limits the utility of hand-crafted features which were utilised in early approaches~\cite{mamonov2014automated,tajbakhsh2015automated}. Deep learning-based approaches for polyp segmentation~\cite{yu2016integrating,zhang2018polyp} have helped to overcome these limitations. Brandao \textit{et al.}~\cite{brandao2017fully} adapt convolution neural networks (FCN) fully by converting three established networks into a full convolution architecture and fine-tuning their learned representations to the polyp segmentation task. Extensions of the original U-Net model~\cite{ronneberger2015u} such as U-Net++~\cite{zhou2018unet++}, ResUNet~\cite{jha2019resunet++} and Dense U-Net~\cite{wang2019dense} were applied to the polyp segmentation task and demonstrated improved accuracy. PsiNet~\cite{murugesan2019psi} focused on region boundary constraints, exploiting polyp region and boundary information. In addition, Fang \textit{et al.}~\cite{fang2019selective} proposed an aggregated multi-scale and multi-sensory field feature aggregation network for polyp segmentation and considered the dependence between region and boundary branches through boundary-sensitive loss functions. PraNet~\cite{fan2020pranet} achieves cutting-edge segmentation accuracy and generalisation performance by mining the relationship between regions and boundary cues through parallel reverse attention modules. The method demonstrated an improvement with mDice over 8\% and 6\% on the Kvasir and CVC-ClinicDB datasets, respectively.
Polyp-PVT~\cite{dong2021polyp} adopted the transformer encoder, which can learn more robust representations. The three modules incorporated in the method are respectively employed for gathering semantic and positional information of polyps from high-level features, capturing polyp information features in the lower layers, and extending pixel features of polyp regions with semantic positional information, which facilitated a more effective feature fusion.
Ta \textit{et al.}~\cite{ta2023ble} devised a boundary learning and enhancement network that integrates edge details into higher-level features through a fusion approach, thereby generating features for segmentation with boundary information. Subsequently, multi-level boundary-aware attention blocks provided ambiguous edge information and enhanced the previous segmentation results.
Qiu \textit{et al.}~\cite{qiu2022bdg} proposed an encoder-decoder network called Boundary Distribution Guided Network (BDG-Net). In the encoder, Gaussian distribution is employed to further extract information about the boundaries of polyps, which is then fused with low-level and high-level features. However, due to its heavy reliance on boundary features, this method exhibits sub-optimal segmentation results when polyps have weak edges.
\cite{lu2023multi} also focuses on polyp boundary information. They proposed a multi-scale global perception approach for aggregating multi-scale information and expanding the receptive field to acquire the primary distribution of polyps at both local and global levels. Experimental results in the paper indicated that their proposed method outperforms most state-of-the-art models on five publicly polyp datasets.
CoInNet~\cite{jain2023coinnet} proposed by Jain \textit{et al.} focused more on tiny polyps. They designed a novel feature extraction strategy that utilizes statistical information between pixel intensities on the feature map. This information was then input into convolutional layers to highlight the polyp regions in the image. The approach achieved mDice of 0.926 and 0.930 on Kvasir and CVC-ClinicDB datasets, which reached state-of-the-art levels.

\subsection{Video polyp segmentation}
Zhao \textit{et al.}~\cite{zhao2022semi} adopted a temporal local context attention mechanism to refine the prediction of the current frame through the prediction results of nearby frames, and a neighbouring frame spatiotemporal attention mechanism used to capture long-range dependencies in time and space. The experimental results in the paper showed that the semi-supervised method they proposed achieves similar results to the fully supervised method. Self-attentive networks~\cite{wang2018non} are able to capture the internal relevance of features and reduce the dependence on external information. Ji \textit{et al.}~\cite{ji2021progressively} proposed PNS-Net using self-attention blocks, equipping with recurrence and CNNs to improve polyp segmentation accuracy and achieve real-time inference speed. They further extended their approach~\cite{ji2022video} focused on video polyp segmentation, which accomplished maximum Dice index of 0.756 and 0.737 in SUN-SEG-Easy and SUN-SEG-Hard. The recently proposed ESFPNet~\cite{chang2023esfpnet} utilized a pre-trained Mix Transformer (MiT) encoder and an efficient staged feature pyramid (ESFP) decoder structure to achieve accurate image segmentation. Although the method focused on bronchial lesion segmentation, it achieved excellent results on the polyp segmentation dataset. The approach obtained mDice of 0.811 and 0.823 in CVC-ColonDB and ETIS-LaribPolypDB, respectively.

\subsection{Self-supervised learning}
%
Self-supervised learning (SSL)~\cite{jaiswal2020survey} has shown its advantages in limited annotated datasets. 
Wang \textit{et al.}~\cite{wang2023foundation} proposed Endo-FM, which contained video converters for capturing local and global long-range dependencies across spatial and temporal dimensions. The Transformer model was then pre-trained using global and local views in a self-supervised manner. 
The paper findings indicated that compared with the SOTA SSL method, it improved 3.1\% F1 score, 4.8\% mDice and 5.5\% F1 score on classification, segmentation and detection, respectively. Debayan \textit{et al.}~\cite{bhattacharya2021self} used training a U-Net to repair randomly missing pixels in an image as a proxy task to obtain a pre-trained model and then fine-tune the model to achieve good performance in a downstream polyp segmentation task. However, most of these methods focus on the segmentation of still images and ignore the temporal information present in colonoscopy videos. This results in a substantial drop in their performance on real colonoscopy video data. 
\begin{figure*}[t!]
    \centering
    \includegraphics[width=1.0\textwidth]{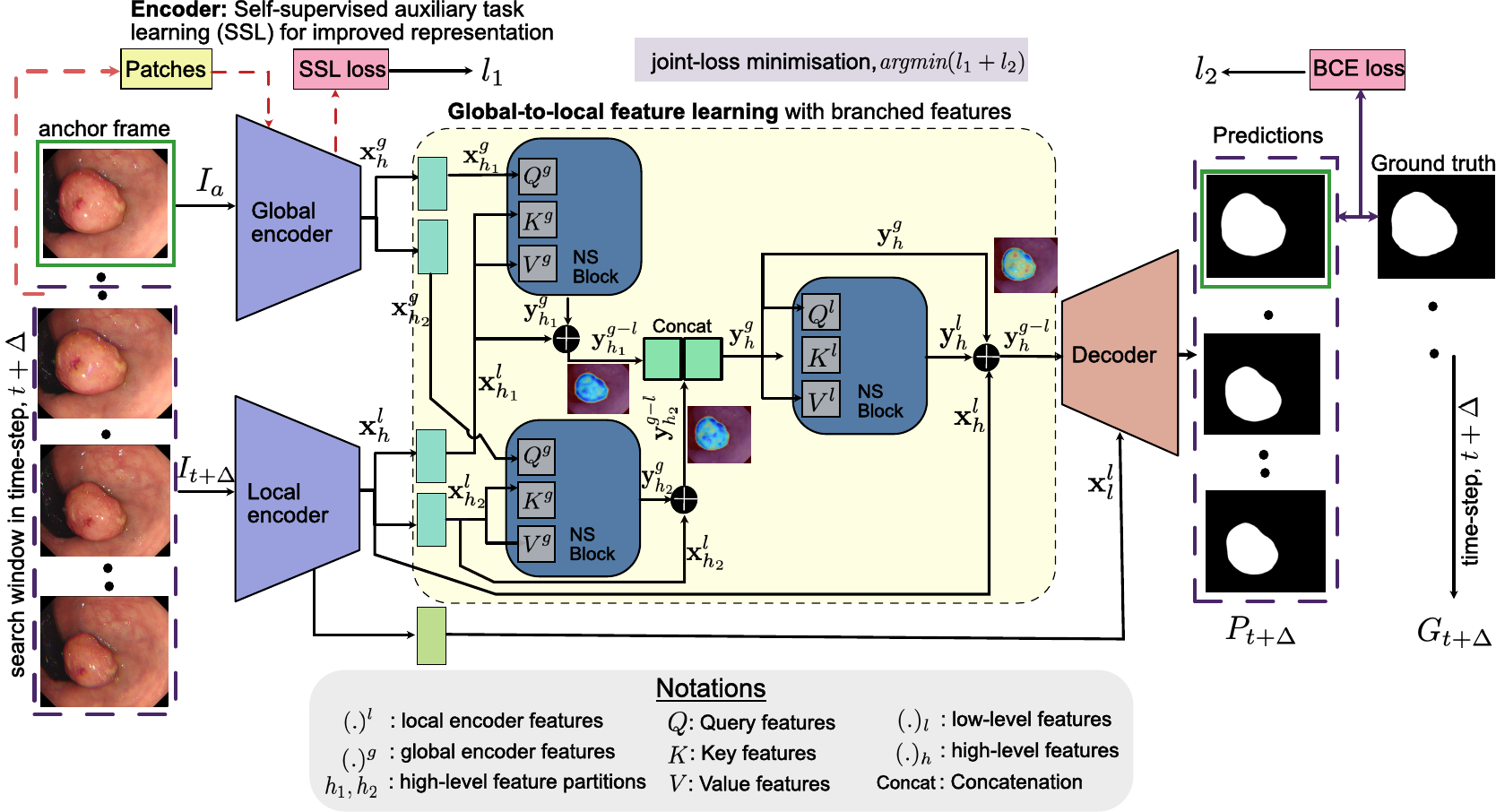}
    \vspace{-3mm}
    \caption{\textbf{Proposed SSTFB network:} Our network consists of three parts: a) Global and local encoders trained with a self-supervision loss comprising Jigsaw-puzzle sampling. b) High-level global and local features branch from both encoders and pass through a normalised self-attention block (NS-block~\cite{ji2021progressively}), enabling global-to-local feature learning. c) A decoder layer that fuses the self-attention high-level feature maps from NS-blocks and the low-level features from the local encoder for final mask prediction. }
    \label{fig:blockdiagram}
\end{figure*}
\begin{figure}[t!]
    \centering
    \includegraphics[width=0.5\textwidth]{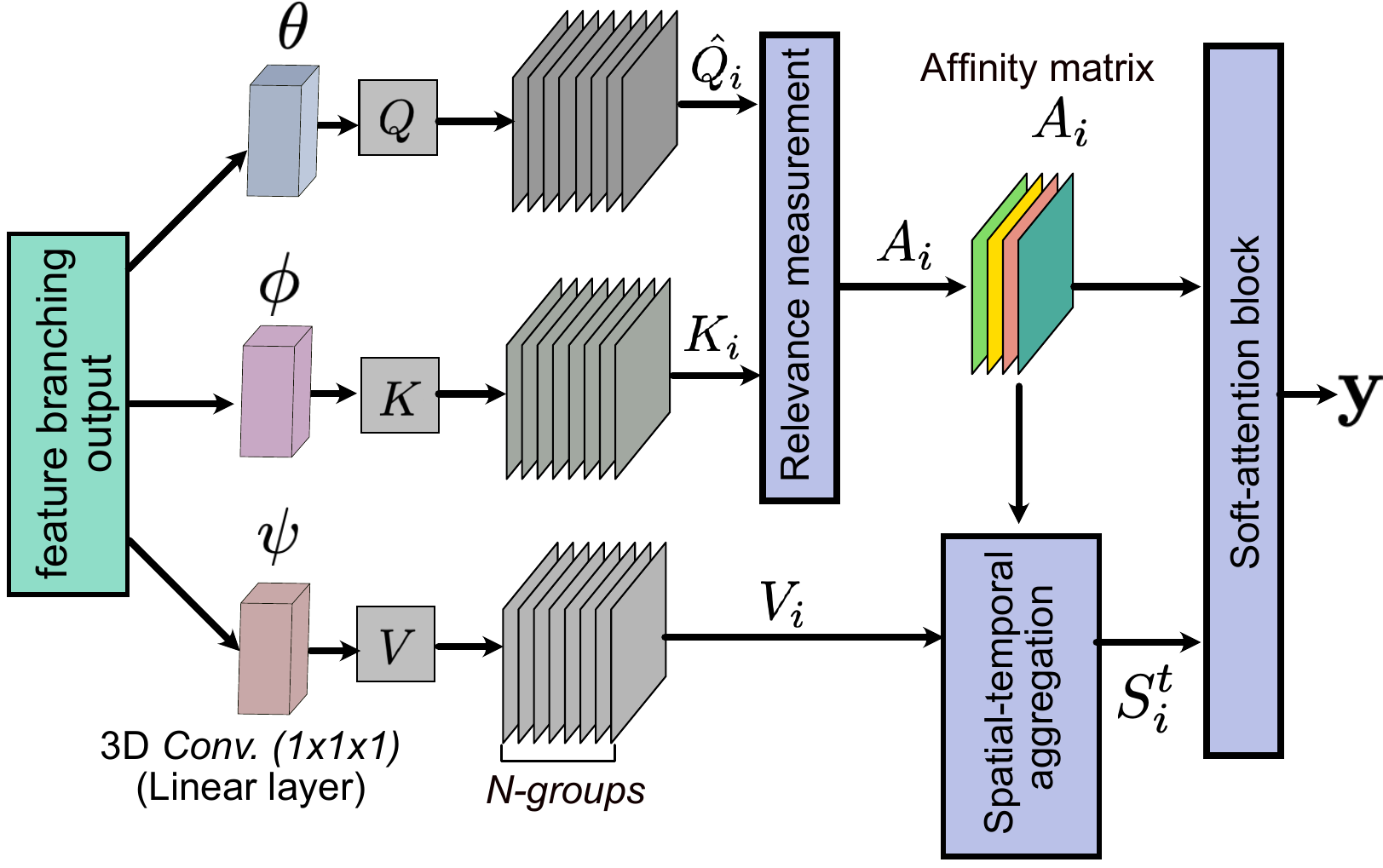}
    \vspace{-3mm}
    \caption{Normalised self-attention block \textbf{(NS-block)} used in our SSTFB network.}
    \label{fig:ns-block}
\end{figure}

\subsection{Our approach}
 In this work, we propose a novel end-to-end self-supervised and normalised self-attention blocks~\cite{ji2021progressively} with optimal configurations and feature branching adhering to real-time performance, namely SSTFB network, demonstrating significant improvement over all state-of-the-art (SOTA) methods. 
Our contributions in this work can be summarised as:
\begin{itemize}
    \item We propose a self-supervised learning scheme as an auxiliary task in an end-to-end framework to boost feature representations.
    \item We use feature branching and concatenation techniques for learning separately semantically meaningful features.
    \item We exploit the spatial-temporal relationship of the features by normalised self-attention blocks with feature aggregation and relevance measurement.
    \item We show optimal choices of normalised self-attention blocks for global-to-local feature learning without impacting real-time performance.
    \item We compare on both seen and unseen video datasets with several SOTA methods where our method outperforms them by a larger margin. 
\end{itemize}
\section{Method}
\subsection{Encoder self-supervision}
We use a pretext-invariant representation learning (PIRL)~\cite{misra2020self} that uses a Jigsaw puzzle~\cite{noroozi2016unsupervised} as a pretext task for representation learning. The encoder is fed with the frame sequence ($\Delta = 5$), which is then used to create patches and shuffled for an auxiliary self-supervised learning (SSL) task. We minimise SSL loss given by a joint function comprising of widely used contrastive loss aimed at exploiting positive and negative samples and a patch-level instance group feature discriminator loss that maximises the inter-separation and minimises intra-separation between patches, enabling learning more salient features~\cite{xu2022patch}. The same encoder is used for generating both global features ($\textbf{x}_h^g$) and local features ($\textbf{x}_h^l$, $\textbf{x}_l^l$).

\subsection{Normalised self-attention block}\label{sec:NS-Block}
A normalised self-attention block as proposed in PNSNet~\cite{ji2021progressively} is used. However, unlike~\cite{ji2021progressively} we use the feature branches as input to the NS-block, \textit{i.e.} each NS-block receives input $\textbf{x}\in \mathcal{R}^{T\times H/2\times W/2 \times C}$ with temporal frames $T$, height $H$, width $W$, and channel $C$ (See Fig.~\ref{fig:ns-block}). Three linear embedding functions $\theta(.)$,  $\phi(.)$, and $\psi(.)$ are used to generate corresponding attention features using a $1\times1\times 1$ convolutional layer corresponding to query $\mathit{Q}$, key $\mathit{K}$ and value $\mathit{V}$. Each attention feature is then divided into $N$ groups, and the query features are further normalised for internal covariance shift compensation by utilising layer normalisation. Thus, the set now can be represented as $\{\hat{Q}_i, K_i, V_i\} \in \mathcal{R}^{T\times H/2\times W/2 \times C/N}$ with $i = \{1, .., N\}$. Similar to ~\cite{ji2021progressively}, we used a query-dependent rule where each query pixel $\textbf{x}^q$ of $\hat{Q}_i$ we require to find corresponding key $K_i$ in a constrained neighbouring window of kernel size $k$ and dilation rate $d_i = 2i-1$ using a sampling function $\mathcal{F}_s$. Finally, an affinity matrix $A_i \in \mathcal{R}^{THW/4 \times T(2k+1)^2}$ is computed as:
\begin{equation}
    A_i = \text{Softmax}(\frac{\hat{Q}_i\mathcal{F}_s<\textbf{x}^q, K_i>^T}{\sqrt{C/N}}. 
\end{equation}

\noindent{For} $\textbf{x}^a\in A_i$ we compute inner product with the value $V_i$ utilising the similar approach as done above with sampling function $\mathcal{F}_s$ and neighbouring window resulting in a spatial-temporal aggregation matrix, $S^T_i = A_i \mathcal{F}_s <\textbf{x}^a, V_i>$. Finally, a soft-attention is applied to get the final output of the block by concatenating group features along the channel and taking maximum value along these channels for both $A_i$ and $S^T_i$.

\subsection{Global-to-local feature learning}
%
Our proposed global-to-local feature learning is unique. Here, we use three NS-blocks exploiting both feature branches from the global encoder for anchor frame $I_a$ and temporal feature information $I_{t+\Delta}$ coming from the local encoder. 
The global encoder produces features for $I_a$ extracted from $\text{conv}4\_6$ layer of Res2Net-50\cite{gao2019res2net}. Features extracted for the local encoder for consecutive frames $I_{t+\Delta}$ is extracted from $\text{conv}3\_4$ and $\text{conv}4\_6$ which are indicated as low-level $(.)_l$ and high-level $(.)_h$ features, respectively.
%
Order of feature maps in the branches of high-level features remains the same while entering to two first NS-blocks, i.e., the top feature branch from global encoder $\textbf{x}^g_{h_1}$ and the top feature branch from local encoders $\textbf{x}^l_{h_1}$ entering the first NS-block while $\textbf{x}^g_{h_2}$ and $\textbf{x}^l_{h_2}$ are fed to the second NS-block.
For arbitrary temporal distance $\Delta$, the first two NS-blocks model the long-term relationship, as $\{\textbf{x}_l^{h_1},\textbf{x}_l^{h_2}\}\in \mathcal{R}^{\Delta \times H/2 \times W/2 \times C}$. Anchor features are taken as query entries while $\textbf{x}_l^{h_1}$ and $\textbf{x}_l^{h_2}$ are taken as key and value features. As discussed in Section~\ref{sec:NS-Block}, NS-block creates a spatial-temporal aggregation and pixel-wise similarities between query and key within a neighbour window such that high-level short-term features are exploited. Finally, we use an element-wise addition of features as a residual operation in both NS-blocks, giving $y_{h_1}^{g-l}$ and $y_{h_2}^{g-l}$, respectively. We then concatenate both features from the NS blocks, which is then passed to a third NS block that further enhances the features to propagate into a local neighbourhood, boosting more local representations and two residual connections to avoid the vanishing gradient problem. It can be observed that for $y_h^{g-l}$, both local and global features are pronounced after performing these operations and hold spatial-temporal information.
\subsection{Decoder}
Learned features from the global-to-local feature block that consists of the high-level spatial-temporal features $y_h^{g-l}$ together with the low-level features from the local encoder for time frame $t$ represented as $\textbf{x}_l^l$ is fed to a two-stage UNet-like decoder $D$ estimating predictions $P_t$. These predictions are compared with ground truth using widely used binary cross-entropy $\mathcal{L}_{\text{BCE}}$.
\subsection{Joint-loss optimisation (end-to-end framework)}
%
Our novel approach includes joint-loss optimisation for both auxiliary tasks, i.e., self-supervised pretext learning (SSL) and the main fully supervised segmentation task, i.e., segmentation in an end-to-end fashion. To this end, we define the SSL loss as the combination of noise contrastive estimator (NCE) and patch-level instance group discriminator loss (PLD)~\cite{xu2022patch}. The NCE loss is defined as:

\begin{equation}{\label{eq:nce}}
\begin{split}
    \mathcal{L}_{NCE}(\textbf{I},\textbf{I}_t) = 0.5 \mathcal{L}_{{NCE}_{I}}(\textbf{m}_{\textbf{I}},  \bar{r}(\phi_\theta(\textbf{I}))) + \\ \nonumber
    0.5 \mathcal{L}_{{NCE}_{{I}_t}}(\textbf{m}_{\textbf{I}}, \bar{r_{t}}(\phi_\theta(\textbf{I}_t)))\\ 
    = 0.5 \{-\log [{h}(\bar{r}(\phi_\theta(\textbf{I})), \textbf{m}_{\textbf{I}})]    \nonumber  \\
   -\sum_{\textbf{I}'\in \mathcal{D}_n} \log [1- {h}( \textbf{m}_{\textbf{I}^{'}}, \bar{r}(\phi_\theta(\textbf{I})))]\} \\
   + 0.5 \{-\log [{h}(\bar{r_{t}}(\phi_\theta(\textbf{I}_t)), \textbf{m}_{\textbf{I}})]    \nonumber  \\
   -\sum_{\textbf{I}'\in \mathcal{D}_n} \log [1- {h}( \textbf{m}_{\textbf{I}^{'}}, \bar{r_{t}}(\phi_\theta(\textbf{I}_t)))]\}.
\end{split}  
\end{equation}

$\textbf{I}$ and $\textbf{I}_t$ represent image samples ($N$ in total) and corresponding transformed image patches respectively. Negative samples $\textbf{I}'$ which are used for measuring the similarity between two representations. An encoder with free parameters $\theta$ is used to incarnate representation $\phi_\theta(.)$. Two fully connected layers, $r(.)$ and $r_{t}(.)$, are included in a unique projection head re-scale positive samples and transformed sample patch representations to the same dimension. $\mathcal{D}_n$ refers to the negative samples list which will increase with the training epochs. 

Memory bank strategy stores all representations $\textbf{m}_\textbf{I}$ from image $\textbf{I}$, which are updated in a moving average manner after each training epoch. Simultaneously, these embedding weights can help to generalise negative samples as $\textbf{m}_{\textbf{I}^{'}}$. 

The noise contrastive estimator is computed between positive samples and corresponding negative samples. $S(.,.)$ computes the $cosine$ similarity between the different level representations, which can be adjusted by a temperature parameter $\tau$. We let the $r_{\textbf{I}}$ as the positive sample representations, $r_{\textbf{I}}^{+}$ as the embedding from $\textbf{m}_\textbf{I}$ and $r_{\textbf{I}}^{-}$ as the negative sample representations from $\textbf{m}_{\textbf{I}^{'}}$.
The posterior probability $h(.,.)$ of the NCE loss can be written as:   

\begin{equation}{\label{eq:cosine}}
    h(r_{\textbf{I}},r_{\textbf{I}}^{+}) = \frac{\exp{(S(r_{\textbf{I}},r_{\textbf{I}}^{+})}/\tau)}{\exp{(S(r_{\textbf{I}},r_{\textbf{I}}^{+})}/\tau) + {\sum_{r_{\textbf{I}}^{-}\in \mathcal{D}_n} \exp{(S(r_{\textbf{I}}, r_{\textbf{I}}^{-})}/\tau)}
    }.
\end{equation}

The limitation in SSL is that if training samples are similar, the negative examples used in the NCE loss are more likely to be composed of highly similar instances. The grouping strategy can help mitigate such limitations. The idea of PLD loss is using cross-entropy loss to compute $cosine$ similarity between normalised image embedding $\bar r(.)$ and patch embedding $r_{t}(.)$, image embedding $r(.)$ and normalised patch embedding $\bar r_{t}(.)$. k-means clustering is used to group the feature vectors and then compute the cluster centroids for each image embedding $C^{k}$ and patch embedding $PC^{k}$ where $k$ is the number of classes. By assuming the group in the patch-level cluster is the same as the group in the image-level for that particular class, therefore the PLD loss is defined as:

\begin{equation}{\label{eq:PLD_loss}}
\begin{split}
\mathcal{L}_{PLD}(\textbf{I},\textbf{I}_t) = 0.5\sum_{\forall k} h(\bar{r}(\phi_\theta(\textbf{I})), PC^k) \\ + 0.5
\sum_{\forall k} h(\bar{r_{t}}(\phi_\theta(\textbf{I}_t)), C^k).
\end{split}
\end{equation}

Finally, for the segmentation task, we utilise predicted segmentation mask $P_t$ and ground-truth mask $G_t$ at the same time frame $t$ to minimise a binary cross-entropy (BCE) loss formulated as:

\begin{equation}{\label{eq:PLD_loss}}
\mathcal{L}_{BCE} = - \sum(G_t \log (P_t)+(1-G_t) \log (1-P_t))
\end{equation}

Thus, the joint-loss $\mathcal{L_{\text{joint}}}$ is minimised by a combination of two equally weighted representation learning SSL approaches, $ \lambda_1 = \lambda_2 = 0.25$. A higher weight is given to the BCE loss for the target segmentation task, $\lambda_3=0.5$. The final loss $\mathcal{L_{\text{joint}}}$ is given by: 
\begin{equation}
    \mathcal{L_{\text{joint}}} = \lambda_1\mathcal{L_\text{NCE}} + \lambda_2\mathcal{L_\text{PLD}} + \lambda_3\mathcal{L}_{\text{BCE}}.
\end{equation}

\section{Experimental setup}
\subsection{Datasets and Comparison models}
Experiments were conducted on two widely used video polyp segmentation datasets: \href{https://www.kaggle.com/datasets/balraj98/cvcclinicdb}{CVC-ClinicVideoDB/CVC-612}~\cite{bernal2015wm}
 and \href{http://amed8k.sundatabase.org/}{SUN-SEG}~\cite{misawa2021development}. 
 SUN-SEG is a large colonoscopy video database from Showa University and Nagoya University. The dataset comprises 49,136 polyp frames extracted from 100 polyp videos and the corresponding masks. We split 40\% of SUN-SEG data for training and the rest of the data for testing. The test data are divided into SUN-SEG-Easy (seen \& unseen) and SUN-SEG-Hard (seen \& unseen) according to the level of difficulty of the polyp pathology. Seen indicates that the testing samples are from the same case in the training set, whereas the unseen denotes that samples do not exist in the training set. We compare our approach with six most recent polyp segmentation methods: PraNet~\cite{fan2020pranet}, Polyp-PVT~\cite{dong2021polyp}, SANet~\cite{wei2021shallow},  PNSNet~\cite{ji2021progressively}, TGANet~\cite{tomar2022tganet} and PNS+~\cite{ji2022video}.
 %
\begin{table*}[t!]
\centering
\scriptsize{
\caption{Quantitative results on seen test datasets \label{tab:test-seen}}
\begin{tabular}{l|l|c|c|c|c|c|c|c|c|c|c}
\hline
\textbf{Method}                    & \textbf{Publication} 
               & \multicolumn{5}{c}{\textbf{SUN-SEG-Easy (seen)}} & \multicolumn{5}{|c}{\textbf{SUN-SEG-Hard (seen)}}\\ 
           \cline{3-12}  &   & \multicolumn{1}{c|}{{mDice}}  & \multicolumn{1}{c|}{{mIoU}}  & \multicolumn{1}{c|}{{$S_\alpha$}} & \multicolumn{1}{c|}{{$E_\phi$}} & \textbf{$F^w_\beta$}  &  \multicolumn{1}{c|}{{mDice}}  & \multicolumn{1}{c|}{{mIoU}}  & \multicolumn{1}{c|}{{$S_\alpha$}} & \multicolumn{1}{c|}{{$E_\phi$}} & \textbf{$F^w_\beta$}  \\ \hline \hline
\multicolumn{1}{l|}{PraNet~\cite{fan2020pranet}}            & \multicolumn{1}{l|}{MICCAI$_{20}$}          & \multicolumn{1}{c|}{0.812}          & \multicolumn{1}{c|}{0.737}          & \multicolumn{1}{c|}{0.857}            & \multicolumn{1}{c|}{0.894}               & \multicolumn{1}{c|}{0.743}  & \multicolumn{1}{c|}{0.795}          & \multicolumn{1}{c|}{0.721}          & \multicolumn{1}{c|}{0.834}            & \multicolumn{1}{c|}{0.860}               & 0.736                  \\ \hline
\multicolumn{1}{l|}{Polyp-PVT~\cite{dong2021polyp}}                               & {\begin{tabular}[t]{@{}l@{}}MICCAI$_{21}$ \end{tabular}}       & \multicolumn{1}{c|}{0.819}      &\multicolumn{1}{c|}{0.740}        & \multicolumn{1}{c|}{0.862}            & \multicolumn{1}{c|}{0.899}            & \multicolumn{1}{c|}{0.748} & \multicolumn{1}{c|}{0.801}        & \multicolumn{1}{c|}{0.727}          & \multicolumn{1}{c|}{0.841}            & \multicolumn{1}{c|}{0.862}            & \multicolumn{1}{c}{0.747}             \\ \hline
\multicolumn{1}{l|}{SANet~\cite{wei2021shallow}}  & \multicolumn{1}{l|}{MICCAI$_{21}$}          & \multicolumn{1}{c|}{0.827}          & \multicolumn{1}{c|}{0.743}         & \multicolumn{1}{c|}{0.865}            & \multicolumn{1}{c|}{0.909}               & \multicolumn{1}{c|}{0.751}      & \multicolumn{1}{c|}{0.807}          & \multicolumn{1}{c|}{0.732}         & \multicolumn{1}{c|}{0.845}            & \multicolumn{1}{c|}{0.866}               & 0.751                 \\ \hline
\multicolumn{1}{l|}{TGANet~\cite{tomar2022tganet}}                               & {\begin{tabular}[t]{@{}l@{}}MICCAI$_{22}$ \end{tabular}}   &\multicolumn{1}{c|}{0.833}          & \multicolumn{1}{c|}{0.752}          & \multicolumn{1}{c|}{0.876}            & \multicolumn{1}{c|}{0.915}              & \multicolumn{1}{c|}{0.76}    & \multicolumn{1}{c|}{0.814}          & \multicolumn{1}{c|}{0.739}          & \multicolumn{1}{c|}{0.851}            & \multicolumn{1}{c|}{0.871}               & \multicolumn{1}{c}{0.755}                        \\ \hline
\multicolumn{1}{l|}{PNSNet~\cite{ji2021progressively}}        & \multicolumn{1}{l|}{MICCAI$_{21}$}          & \multicolumn{1}{c|}{0.835}          & \multicolumn{1}{c|}{0.760}          & \multicolumn{1}{c|}{0.882}            & \multicolumn{1}{c|}{0.922}               & \multicolumn{1}{c|}{0.771}      & \multicolumn{1}{c|}{0.821}          & \multicolumn{1}{c|}{0.753}          & \multicolumn{1}{c|}{0.860}            & \multicolumn{1}{c|}{0.882}               & 0.769                   \\ \hline
\multicolumn{1}{l|}{PNS+~\cite{ji2022video}}                               & {\begin{tabular}[t]{@{}l@{}}MIR$_{22}$ \end{tabular}}       & \multicolumn{1}{c|}{0.887}        & \multicolumn{1}{c|}{0.826}          &\multicolumn{1}{c|}{0.915}           & \multicolumn{1}{c|}{0.954}  & \multicolumn{1}{c|}{0.844}   & \multicolumn{1}{c|}{0.855}        & \multicolumn{1}{c|}{0.787} & \multicolumn{1}{c|}{0.885}           &\multicolumn{1}{c|}{0.927}               & \multicolumn{1}{c}{0.796}           \\ \hline
\multicolumn{1}{l|}{CFANet~\cite{zhou2023cross}}                               & {\begin{tabular}[t]{@{}l@{}}PR$_{23}$ \end{tabular}}       & \multicolumn{1}{c|}{0.841}         & \multicolumn{1}{c|}{0.767}        & \multicolumn{1}{c|}{0.88}            & \multicolumn{1}{c|}{0.88}            & \multicolumn{1}{c|}{0.785}   & \multicolumn{1}{c|}{0.829}         & \multicolumn{1}{c|}{0.757}         & \multicolumn{1}{c|}{0.866}           & \multicolumn{1}{c|}{0.866}             & \multicolumn{1}{c}{0.762}            \\ \hline
\multicolumn{1}{l|}{ICBNet~\cite{xiao2022icbnet}}                               & {\begin{tabular}[t]{@{}l@{}}BIBM$_{23}$ \end{tabular}}       & \multicolumn{1}{c|}{0.846}         & \multicolumn{1}{c|}{0.775}         & \multicolumn{1}{c|}{0.887}            & \multicolumn{1}{c|}{0.921}             & \multicolumn{1}{c|}{0.781}  & \multicolumn{1}{c|}{0.827}        & \multicolumn{1}{c|}{0.755}       & \multicolumn{1}{c|}{0.862}           & \multicolumn{1}{c|}{0.887}              & \multicolumn{1}{c}{0.767}            \\ \hline
\multicolumn{1}{l|}{PVT-CASCADE~\cite{rahman2023medical}}                               & {\begin{tabular}[t]{@{}l@{}}WACV$_{23}$ \end{tabular}}       & \multicolumn{1}{c|}{0.855}        & \multicolumn{1}{c|}{0.789}         &\multicolumn{1}{c|}{0.789}           & \multicolumn{1}{c|}{0.93}            & \multicolumn{1}{c|}{0.792}   & \multicolumn{1}{c|}{0.832}         & \multicolumn{1}{c|}{0.762}         & \multicolumn{1}{c|}{0.869}            & \multicolumn{1}{c|}{0.891}            & \multicolumn{1}{c}{0.783}           \\ \hline
\multicolumn{1}{l|}{CTNet~\cite{xiao2024ctnet}}                               & {\begin{tabular}[t]{@{}l@{}}IEEE TOC$_{24}$ \end{tabular}}       & \multicolumn{1}{c|}{0.867}        & \multicolumn{1}{c|}{0.807}         & \multicolumn{1}{c|}{0.899}            & \multicolumn{1}{c|}{0.939}            & \multicolumn{1}{c|}{0.816}  & \multicolumn{1}{c|}{0.84}        & \multicolumn{1}{c|}{0.769}        &  \multicolumn{1}{c|}{0.873}            & \multicolumn{1}{c|}{0.895}              & \multicolumn{1}{c}{0.787}           \\ \hline
\multicolumn{1}{l|}{SSTFB}               & \multicolumn{1}{l|}{Ours$_{24}$}          & \multicolumn{1}{c|}{\textbf{0.911}}          & \multicolumn{1}{c|}{\textbf{0.856}}          & \multicolumn{1}{c|}{\textbf{0.935}}            & \multicolumn{1}{c|}{\textbf{0.968}}              & \textbf{0.871}       & \multicolumn{1}{c|}{\textbf{0.881} }         & \multicolumn{1}{c|}{\textbf{0.813}}          & \multicolumn{1}{c|}{\textbf{0.909}}            & \multicolumn{1}{c|}{\textbf{0.942}}               & \textbf{0.829}         \\ \hline

\hline
\end{tabular}
}
\end{table*}
\subsection{Evaluation metrics}
Two conventional metrics, mean Dice Similarity Coefficient ($ mDice =\frac{2TP}{2TP+FP+FN}$) and mean Intersection over Union ($ mIoU =\frac{TP}{TP+FP+FN}$) to measure the similarity and overlap between predictions and ground truth are used. Here, TP is true positives, FP is false positives, and FN refers to the false negatives. 

In addition, we have also included the S-measure, which is the weighted ($\alpha$) metric between the object-aware structural similarity measure ($S_o$) and region-aware structural similarity measure ($S_r$)~\cite{fan2017structure} ($S_\alpha = \alpha\times S_o(P, G)+(1-\alpha)\times S_r(P, G)$). Here, $P$ is  prediction, $G$ is ground true, and $\alpha$ is set to 0.5. 

A weighted F-measure which corrects for ``Equal-importance flaws'' in Dice is also used~\cite{margolin2014evaluate, achanta2009frequency} ($F^w_\beta=(1+\beta^{2})\times precision^{w} \times recall^{w}/(\beta^{2} \times (precision^{w} + recall^{w}))$. Here, $precison = \frac{TP}{TP+FP}$, $recall = \frac{TP}{TP+FN}$, and $\beta^{2}=0.3$. 

Another metric referred to as an enhanced-alignment measure which is suitable for assessing heterogeneous location and shape of polyps in colonoscopy~\cite{ji2022video} is also used ($E_\phi = (1/(W \times H))\sum_{x=0}^W \sum_{y=0}^H \phi (P(x,y), G(x,y))$, $W$: width of $G$, $H$: height of $G$, $\phi$: enhanced alignment matrix)~\cite{fan2018enhanced}.
\subsection{Implementation details}
We train our SSTFB network using the PyTorch framework on a server deployed with NVIDIA Quadro RTX 6000 graphics cards. All input images were resized to $256\times 448$ pixels, and batch size was set to 24. The first video clip frame was chosen as an anchor and randomly sampled five consecutive frames as neighbouring frames ($\Delta=5$). Throughout the model training process, we set the initial learning rate and weight decay of the Adam optimiser to 3$e^{-4}$ and 1$e^{-4}$, respectively. The number of stacked normalised self-attention blocks is set as $R = 3$, the kernel size $k = 3$, and the number of attention groups to $N = 4$ with dilation rate of $\{3,4,3,4\}$ for the first two NS-blocks, and dilation rate of $\{1,2,1,2\}$ for the third NS-block. For the SSL task, we use the patches from the entire sequence of images fed to the Res2Net-50 model (pre-trained with ImageNet weights).\\

\section{Results}

\subsection{SOTA method comparisons}
In this section, we assess the performance of our method on two seen test (held-out) datasets (SUN-SEG-Easy(seen) \& SUN-SEG-Hard(seen)) and comparison with other SOTA methods.\\
\subsubsection{Quantitative results}
Quantitative results of SOTA methods and our proposed SSTFB are summarised in Table~\ref{tab:test-seen}. It can be observed from Table~\ref{tab:test-seen} that the performance of the method developed for still polyp image segmentation (PraNet, Polyp-PVT, SANet, TGANet, CFANet, PVT-CASCADE and CTNet) is lower than the video polyp image segmentation method (PNSNet, PNS+ ,ICBNet and SSTFB(ours)) on both SUN-SEG-Easy and SUN-SEG-Hard datasets. Specifically, our proposed SSTFB reaches 0.911 on mDice, 0.856 on mIoU, $S_\alpha$ on 0.935, $E_\phi$ on 0.968 and $F^w_\beta$ on 0.871 achieving performance improvements of 9.9\%, 11.9\%, 7.8\%, 7.4\% and 12.8\% respectively compared with the polyp segmentation benchmark work (PraNet). {SSTFB also achieves 4.4\% and 4.9\% increase of mDice and mIoU on SUN-SEG-Easy(seen) when contrasted with the recent SOTA method (CTNet)}. Simultaneously, our method still has a mDice improvement of about 3\% for the most recent SOTA polyp video segmentation model PNS+. On the SUN-SEG-Hard(seen) dataset, SSTFB presents mDice of 0.881 and mIoU of 0.813, which shows an over 8\% performance improvement compared with PraNet and 2.6\% improvement compared with PNS+.\\
Table~\ref{tab:fps} shows SSTFB can run at extremely high inference speeds (126fps). Our proposed SSTFB significantly outperforms all other SOTA models on all metrics with real-time performance. This shows that our model has a stronger learning ability and can effectively learn semantically meaningful features for polyp segmentation. \\
\begin{table}[t!h!]
\centering
\caption{ Trainable parameters and frames-per-second (FPS)\label{tab:fps}}
\begin{tabular}{l|c|c}
\hline
\multicolumn{1}{l|}{\textbf{Method}}                    & \multicolumn{1}{c|}{\textbf{Trainable parameters}} & \multicolumn{1}{c}{\textbf{FPS}}  \\ \hline\hline
\multicolumn{1}{l|}{PraNet}                               & {\begin{tabular}[t]{@{}l@{}}39.2M	 \end{tabular}}       & \multicolumn{1}{c}{97fps}             \\ \hline
\multicolumn{1}{l|}{Polyp-PVT}                               & {\begin{tabular}[t]{@{}l@{}}125.6M		 \end{tabular}}       & \multicolumn{1}{c}{66fps}  \\ \hline
\multicolumn{1}{l|}{SANet}                               & {\begin{tabular}[t]{@{}l@{}}45.1M			 \end{tabular}}       & \multicolumn{1}{c}{72fps}  \\ \hline
\multicolumn{1}{l|}{TGANet}                               & {\begin{tabular}[t]{@{}l@{}}42.3M			 \end{tabular}}       & \multicolumn{1}{c}{85fps}  \\ \hline
\multicolumn{1}{l|}{PNSNet}                               & {\begin{tabular}[t]{@{}l@{}}29.7M			 \end{tabular}}       & \multicolumn{1}{c}{135fps}  \\ \hline
\multicolumn{1}{l|}{PNS+}                               & {\begin{tabular}[t]{@{}l@{}}26.8M			 \end{tabular}}       & \multicolumn{1}{c}{140fps}  \\ \hline
\multicolumn{1}{l|}{CFANet}                               & {\begin{tabular}[t]{@{}l@{}}25.2M			 \end{tabular}}       & \multicolumn{1}{c}{86fps}  \\ \hline
\multicolumn{1}{l|}{ICBNet}                               & {\begin{tabular}[t]{@{}l@{}}62.1M		 \end{tabular}}       & \multicolumn{1}{c}{71fps} \\ \hline
\multicolumn{1}{l|}{PVT-CASCADE}                               & {\begin{tabular}[t]{@{}l@{}}134.5M			 \end{tabular}}       & \multicolumn{1}{c}{62fps}  \\ \hline
\multicolumn{1}{l|}{CTNet}                               & {\begin{tabular}[t]{@{}l@{}}168.9M			 \end{tabular}}       & \multicolumn{1}{c}{57fps}  \\ \hline
\multicolumn{1}{l|}{SSTFB (ours)}                               & {\begin{tabular}[t]{@{}l@{}}33.4M			 \end{tabular}}       &\multicolumn{1}{c}{126fps}\\ \hline
\end{tabular}
\end{table}

\subsection{Generalization}
For the generalisability study, we have trained our method and other SOTA methods on the SUN-SEG dataset and experimented on two new unseen and one out-of-centre polyp video datasets, namely, SUN-SEG-Easy(unseen), SUN-SEG-Hard(unseen) and CVC-612.\\

\subsubsection{Generalizability results on unseen datasets}
{As can be seen from Table~\ref{tab:test-unseeen}, our method is consistently superior to all SOTA methods on SUN-SEG-Easy(unseen) dataset(e.g., nearly 5\%-10\% higher value on mDice and other metrics compared to PraNet, Polyp-PVT, SANet,  TGANet, CFANet, ICBNet, PVT-CASCADE and CTNet).} Compared to SOTA video polyp segmentation methods (PNS+), our approach still shows over 3\% improvement in mDice, mIou and $S_\alpha$. Especially in SUN-SEG-Hard(unseen) dataset, SSTFB still maintains competitive performance of 0.767, 0.673, 0.822, 0.843 and 0.728 on mDice, mIoU, $S_\alpha$, $E_\phi$ and $F^w_\beta$, respectively. The performance of most methods drops sharply while our method retains its performance with performance improvement even over 10\% compared to most SOTA methods.\\

\subsubsection{Generalizability results on out-of-centre datasets}
We also evaluated our method against out-of-centre data in Table~\ref{tab:test-oos} where we can observe that our proposed method achieves a mDice of 0.862, mIoU of 0.786, $S_\alpha$ of 0.893, $E_\phi$ of 0.923 and $F^w_\beta$ of 0.814, respectively. Compared to PraNet and PNS+, our approach exhibits over 4\% and 2\% enhancement in mDice. Overall, the proposed SSTFB achieves superior performance compared to all other SOTA methods. 

\begin{table*}[t!]
\centering
\scriptsize{
\caption{Quantitative results on unseen test datasets\label{tab:test-unseeen}}
\begin{tabular}{l|l|c|c|c|c|c|c|c|c|c|c}
\hline
\textbf{Method}                    & \textbf{Publication} 
               & \multicolumn{5}{c}{\textbf{SUN-SEG-Easy (unseen)}} & \multicolumn{5}{|c}{\textbf{SUN-SEG-Hard (unseen)}}\\ 
           \cline{3-12}  &   & \multicolumn{1}{c|}{{mDice}}  & \multicolumn{1}{c|}{{mIoU}}  & \multicolumn{1}{c|}{{$S_\alpha$}} & \multicolumn{1}{c|}{{$E_\phi$}} & \textbf{$F^w_\beta$}  &  \multicolumn{1}{c|}{{mDice}}  & \multicolumn{1}{c|}{{mIoU}}  & \multicolumn{1}{c|}{{$S_\alpha$}} & \multicolumn{1}{c|}{{$E_\phi$}} & \textbf{$F^w_\beta$}  \\ \hline\hline
\multicolumn{1}{l|}{PraNet~\cite{fan2020pranet}}            & \multicolumn{1}{l|}{\scriptsize{MICCAI$_{20}$}}         & \multicolumn{1}{c|}{0.669}          & \multicolumn{1}{c|}{0.581}          & \multicolumn{1}{c|}{0.727}            & \multicolumn{1}{c|}{0.754}               & 0.629        & \multicolumn{1}{c|}{0.651}          & \multicolumn{1}{c|}{0.572}          & \multicolumn{1}{c|}{0.719}            & \multicolumn{1}{c|}{0.741}               & 0.601                            \\ \hline
\multicolumn{1}{l|}{Polyp-PVT~\cite{dong2021polyp}}                               & \scriptsize{MICCAI$_{21}$}       &\multicolumn{1}{c|}{0.675}          & \multicolumn{1}{c|}{0.589}         & \multicolumn{1}{c|}{0.737}            & \multicolumn{1}{c|}{0.760}             & \multicolumn{1}{c|}{0.641} & \multicolumn{1}{c|}{0.663}          & \multicolumn{1}{c|}{0.592}         & \multicolumn{1}{c|}{0.706}            & \multicolumn{1}{c|}{0.737}              & \multicolumn{1}{c}{0.593}               \\ \hline
\multicolumn{1}{l|}{SANet~\cite{wei2021shallow}}                       & \scriptsize{MICCAI$_{21}$}          & \multicolumn{1}{c|}{0.687}          & \multicolumn{1}{c|}{0.594}         & \multicolumn{1}{c|}{0.723}            & \multicolumn{1}{c|}{0.762}               & \multicolumn{1}{c|}{0.645}       & \multicolumn{1}{c|}{0.654}          & \multicolumn{1}{c|}{0.565}         & \multicolumn{1}{c|}{0.713}            & \multicolumn{1}{c|}{0.745}               & \multicolumn{1}{c}{0.617}         \\ \hline
\multicolumn{1}{l|}{TGANet~\cite{tomar2022tganet}}                               & \scriptsize{MICCAI$_{22}$}      & \multicolumn{1}{c|}{0.672}          & \multicolumn{1}{c|}{0.577}         & \multicolumn{1}{c|}{0.753}           & \multicolumn{1}{c|}{0.759}             &\multicolumn{1}{c|}{0.631}  & \multicolumn{1}{c|}{0.665}         & \multicolumn{1}{c|}{0.577}        & \multicolumn{1}{c|}{0.749}            & \multicolumn{1}{c|}{0.768}              & \multicolumn{1}{c}{0.638}                       \\ \hline
\multicolumn{1}{l|}{PNSNet~\cite{ji2021progressively}}        & \scriptsize{MICCAI$_{21}$}          & \multicolumn{1}{c|}{0.703}          & \multicolumn{1}{c|}{0.619}          & \multicolumn{1}{c|}{0.764}            & \multicolumn{1}{c|}{0.778}               & \multicolumn{1}{c|}{0.653}      & \multicolumn{1}{c|}{0.676}          & \multicolumn{1}{c|}{0.581}          & \multicolumn{1}{c|}{0.765}            & \multicolumn{1}{c|}{0.776}               & 0.651                 \\ \hline
\multicolumn{1}{l|}{PNS+~\cite{ji2022video}}                               & \scriptsize{MIR$_{22}$}       & 
\multicolumn{1}{c|}{0.749}         &\multicolumn{1}{c|}{0.652}         & \multicolumn{1}{c|}{0.793}          & \multicolumn{1}{c|}{0.829}              & \multicolumn{1}{c|}{0.711}   & \multicolumn{1}{c|}{0.736}        & \multicolumn{1}{c|}{0.639}        &  \multicolumn{1}{c|}{0.795}            & \multicolumn{1}{c|}{0.815}             & \multicolumn{1}{c}{0.704}        \\ \hline
\multicolumn{1}{l|}{SSTFB}               & \scriptsize{Ours$_{24}$}       & \multicolumn{1}{c|}{\textbf{0.772}}          & \multicolumn{1}{c|}{\textbf{0.687}}          & \multicolumn{1}{c|}{\textbf{0.846}}            & \multicolumn{1}{c|}{\textbf{0.877}}               & \textbf{0.748}         & \multicolumn{1}{c|}{\textbf{0.767}}          & \multicolumn{1}{c|}{\textbf{0.673}}          & \multicolumn{1}{c|}{\textbf{0.822}}            & \multicolumn{1}{c|}{\textbf{0.843}}               & \textbf{0.728}                            \\ \hline
\hline
\end{tabular}
}
\end{table*}
\begin{table*}[t!]
\centering
\scriptsize{
\caption{Quantitative results on unseen test datasets\label{tab:test-unseeen}}
\begin{tabular}{l|l|c|c|c|c|c|c|c|c|c|c}
\hline
\textbf{Method}                    & \textbf{Publication} 
               & \multicolumn{5}{c}{\textbf{SUN-SEG-Easy (unseen)}} & \multicolumn{5}{|c}{\textbf{SUN-SEG-Hard (unseen)}}\\ 
           \cline{3-12}  &   & \multicolumn{1}{c|}{{mDice}}  & \multicolumn{1}{c|}{{mIoU}}  & \multicolumn{1}{c|}{{$S_\alpha$}} & \multicolumn{1}{c|}{{$E_\phi$}} & \textbf{$F^w_\beta$}  &  \multicolumn{1}{c|}{{mDice}}  & \multicolumn{1}{c|}{{mIoU}}  & \multicolumn{1}{c|}{{$S_\alpha$}} & \multicolumn{1}{c|}{{$E_\phi$}} & \textbf{$F^w_\beta$}  \\ \hline\hline
\multicolumn{1}{l|}{PraNet~\cite{fan2020pranet}}            & \multicolumn{1}{l|}{\scriptsize{MICCAI$_{20}$}}         & \multicolumn{1}{c|}{0.669}          & \multicolumn{1}{c|}{0.581}          & \multicolumn{1}{c|}{0.727}            & \multicolumn{1}{c|}{0.754}               & 0.629        & \multicolumn{1}{c|}{0.651}          & \multicolumn{1}{c|}{0.572}          & \multicolumn{1}{c|}{0.719}            & \multicolumn{1}{c|}{0.741}               & 0.601                            \\ \hline
\multicolumn{1}{l|}{Polyp-PVT~\cite{dong2021polyp}}                               & \scriptsize{MICCAI$_{21}$}       &\multicolumn{1}{c|}{0.675}          & \multicolumn{1}{c|}{0.589}         & \multicolumn{1}{c|}{0.737}            & \multicolumn{1}{c|}{0.760}             & \multicolumn{1}{c|}{0.641} & \multicolumn{1}{c|}{0.663}          & \multicolumn{1}{c|}{0.592}         & \multicolumn{1}{c|}{0.706}            & \multicolumn{1}{c|}{0.737}              & \multicolumn{1}{c}{0.593}               \\ \hline
\multicolumn{1}{l|}{SANet~\cite{wei2021shallow}}                       & \scriptsize{MICCAI$_{21}$}          & \multicolumn{1}{c|}{0.687}          & \multicolumn{1}{c|}{0.594}         & \multicolumn{1}{c|}{0.723}            & \multicolumn{1}{c|}{0.762}               & \multicolumn{1}{c|}{0.645}       & \multicolumn{1}{c|}{0.654}          & \multicolumn{1}{c|}{0.565}         & \multicolumn{1}{c|}{0.713}            & \multicolumn{1}{c|}{0.745}               & \multicolumn{1}{c}{0.617}         \\ \hline
\multicolumn{1}{l|}{TGANet~\cite{tomar2022tganet}}                               & \scriptsize{MICCAI$_{22}$}      & \multicolumn{1}{c|}{0.672}          & \multicolumn{1}{c|}{0.577}         & \multicolumn{1}{c|}{0.753}           & \multicolumn{1}{c|}{0.759}             &\multicolumn{1}{c|}{0.631}  & \multicolumn{1}{c|}{0.665}         & \multicolumn{1}{c|}{0.577}        & \multicolumn{1}{c|}{0.749}            & \multicolumn{1}{c|}{0.768}              & \multicolumn{1}{c}{0.638}                       \\ \hline
\multicolumn{1}{l|}{PNSNet~\cite{ji2021progressively}}        & \scriptsize{MICCAI$_{21}$}          & \multicolumn{1}{c|}{0.703}          & \multicolumn{1}{c|}{0.619}          & \multicolumn{1}{c|}{0.764}            & \multicolumn{1}{c|}{0.778}               & \multicolumn{1}{c|}{0.653}      & \multicolumn{1}{c|}{0.676}          & \multicolumn{1}{c|}{0.581}          & \multicolumn{1}{c|}{0.765}            & \multicolumn{1}{c|}{0.776}               & 0.651                 \\ \hline
\multicolumn{1}{l|}{PNS+~\cite{ji2022video}}                               & \scriptsize{MIR$_{22}$}       & 
\multicolumn{1}{c|}{0.749}         &\multicolumn{1}{c|}{0.652}         & \multicolumn{1}{c|}{0.793}          & \multicolumn{1}{c|}{0.829}              & \multicolumn{1}{c|}{0.711}   & \multicolumn{1}{c|}{0.736}        & \multicolumn{1}{c|}{0.639}        &  \multicolumn{1}{c|}{0.795}            & \multicolumn{1}{c|}{0.815}             & \multicolumn{1}{c}{0.704}        \\ \hline
\multicolumn{1}{l|}{SSTFB}               & \scriptsize{Ours$_{24}$}       & \multicolumn{1}{c|}{\textbf{0.772}}          & \multicolumn{1}{c|}{\textbf{0.687}}          & \multicolumn{1}{c|}{\textbf{0.846}}            & \multicolumn{1}{c|}{\textbf{0.877}}               & \textbf{0.748}         & \multicolumn{1}{c|}{\textbf{0.767}}          & \multicolumn{1}{c|}{\textbf{0.673}}          & \multicolumn{1}{c|}{\textbf{0.822}}            & \multicolumn{1}{c|}{\textbf{0.843}}               & \textbf{0.728}                            \\ \hline
\hline
\end{tabular}
}
\end{table*}

\begin{table*}[t!]
\centering
\scriptsize{
\caption{Quantitative results on out-of-centre test dataset\label{tab:test-oos}}
\begin{tabular}{l|l|c|c|c|c|c}
\hline
\textbf{Method}                    & \textbf{Publication} 
               & \multicolumn{5}{c}{\textbf{CVC-612}} \\ 
           \cline{3-7}  & & \multicolumn{1}{c|}{{mDice}}  & \multicolumn{1}{c|}{{mIoU}}  & \multicolumn{1}{c|}{{$S_\alpha$}} & \multicolumn{1}{c|}{{$E_\phi$}} & \textbf{$F^w_\beta$}  \\ \hline\hline
\multicolumn{1}{l|}{PraNet~\cite{fan2020pranet}}            & \multicolumn{1}{l|}{MICCAI$_{20}$}          & \multicolumn{1}{c|}{0.817}          & \multicolumn{1}{c|}{0.733}          & \multicolumn{1}{c|}{0.856}            & \multicolumn{1}{c|}{0.885}               & 0.752               \\ \hline
\multicolumn{1}{l|}{Polyp-PVT~\cite{dong2021polyp}}                               & {\begin{tabular}[t]{@{}l@{}}MICCAI$_{21}$ \end{tabular}}       & \multicolumn{1}{c|}{0.826}          & \multicolumn{1}{c|}{0.744}          & \multicolumn{1}{c|}{0.861}            & \multicolumn{1}{c|}{0.891}               & \multicolumn{1}{c}{0.769}               \\ \hline
\multicolumn{1}{l|}{SANet~\cite{wei2021shallow}}                       & \multicolumn{1}{l|}{MICCAI$_{21}$}          & \multicolumn{1}{c|}{0.819}          & \multicolumn{1}{c|}{0.738}         & \multicolumn{1}{c|}{0.849}            & \multicolumn{1}{c|}{0.889}               & 0.763               \\ \hline
\multicolumn{1}{l|}{TGANet~\cite{tomar2022tganet}}                               & {\begin{tabular}[t]{@{}l@{}}MICCAI$_{22}$ \end{tabular}}       & \multicolumn{1}{c|}{0.832}         & \multicolumn{1}{c|}{0.751}         & \multicolumn{1}{c|}{0.879}            & \multicolumn{1}{c|}{0.896}             & \multicolumn{1}{c}{0.775}               \\ \hline
\multicolumn{1}{l|}{PNSNet~\cite{ji2021progressively}}        & \multicolumn{1}{l|}{MICCAI$_{21}$}          & \multicolumn{1}{c|}{0.822}          & \multicolumn{1}{c|}{0.739}          & \multicolumn{1}{c|}{0.857}            & \multicolumn{1}{c|}{0.883}               & 0.752               \\ \hline
\multicolumn{1}{l|}{PNS+~\cite{ji2022video}}                               & {\begin{tabular}[t]{@{}l@{}}MIR$_{22}$ \end{tabular}}      & \multicolumn{1}{c|}{0.841}          & \multicolumn{1}{c|}{0.763}          & \multicolumn{1}{c|}{0.882}            & \multicolumn{1}{c|}{0.907}               & \multicolumn{1}{c}{0.793}              \\ \hline
\multicolumn{1}{l|}{SSTFB}               & \multicolumn{1}{l|}{Ours$_{24}$}          & \multicolumn{1}{c|}{\textbf{0.862}}          & \multicolumn{1}{c|}{\textbf{0.786}}          & \multicolumn{1}{c|}{\textbf{0.893}}            & \multicolumn{1}{c|}{\textbf{0.923}}               & \textbf{0.814}               \\ \hline
\end{tabular}
}
\end{table*}
%
\begin{table*}[t!]
\centering
\scriptsize{
\caption{Ablation experiments of SSTFB on different datasets\label{tab:ablation_datasets}}
\begin{tabular}{l|c|c|c|c|c|c|c|c|c|c}
\hline
\textbf{Type}  & \multicolumn{2}{c}{\textbf{Easy(seen)}} & \multicolumn{2}{|c}{\textbf{Hard(seen)}}& \multicolumn{2}{|c}{\textbf{Easy(unseen)}}& \multicolumn{2}{|c}{\textbf{Hard(unseen)}}& \multicolumn{2}{|c}{\textbf{CVC-612}}\\ 
           \cline{2-11}  &  \multicolumn{1}{c|}{{mDice}}  & \multicolumn{1}{c|}{{$S_\alpha$}}  & \multicolumn{1}{c|}{{mDice}} & \multicolumn{1}{c|}{{$S_\alpha$}} & \multicolumn{1}{c|}{mDice}  &  \multicolumn{1}{c|}{{$S_\alpha$}}  & \multicolumn{1}{c|}{{mDice}}  & \multicolumn{1}{c|}{{$S_\alpha$}} & \multicolumn{1}{c|}{{mDice}} & \textbf{$S_\alpha$}  \\ \hline\hline
{3NS-block w/o SSL}        & 0.931        & \multicolumn{1}{c|}{0.937}          & \multicolumn{1}{c|}{0.895}          & \multicolumn{1}{c|}{0.913}            &\multicolumn{1}{c|}{0.822}          & \multicolumn{1}{c|}{0.874}          & \multicolumn{1}{c|}{0.791}            & \multicolumn{1}{c|}{0.839}               &  \multicolumn{1}{c|}{0.877}               & 0.919                            \\ \hline           
{3NS-block + SSL w/o e2e}        & 0.936        & \multicolumn{1}{c|}{0.947}          & \multicolumn{1}{c|}{0.897}          & \multicolumn{1}{c|}{0.921}            &\multicolumn{1}{c|}{0.816}          & \multicolumn{1}{c|}{0.871}          & \multicolumn{1}{c|}{0.806}            & \multicolumn{1}{c|}{0.854}               &  \multicolumn{1}{c|}{0.885}               & 0.927                            \\ \hline
{3NS-block + SSL in e2e}       &  0.943        & \multicolumn{1}{c|}{0.950}          & \multicolumn{1}{c|}{0.902}          & \multicolumn{1}{c|}{0.929}            & \multicolumn{1}{c|}{0.829}          & \multicolumn{1}{c|}{0.878}          & \multicolumn{1}{c|}{0.815}            & \multicolumn{1}{c|}{0.866}               &\multicolumn{1}{c|}{0.903}               & 0.931                            \\ \hline
 {3NS-block + SSL in e2e+b}        &  \textbf{0.959}        & \multicolumn{1}{c|}{\textbf{0.962}}          & \multicolumn{1}{c|}{\textbf{0.925}}          & \multicolumn{1}{c|}{\textbf{0.938}}            &\multicolumn{1}{c|}{\textbf{0.841}}          & \multicolumn{1}{c|}{\textbf{0.899}}          & \multicolumn{1}{c|}{\textbf{0.827}}            & \multicolumn{1}{c|}{\textbf{0.871}}               & \multicolumn{1}{c|}{\textbf{0.911}}               & \textbf{0.942}                            \\ \hline
\end{tabular}
}
\end{table*}

\begin{table*}[t!h!]
\centering
\caption{Ablation with different NS-blocks in SSTFB\label{tab:ablation_ns}}
\begin{tabular}{l|c|c|c|c|c|c|c|c|c|c}
\hline
\textbf{Type}  & \multicolumn{2}{c}{\textbf{Easy(seen)}} & \multicolumn{2}{|c}{\textbf{Hard(seen)}}& \multicolumn{2}{|c}{\textbf{Easy(unseen)}}& \multicolumn{2}{|c}{\textbf{Hard(unseen)}}& \multicolumn{2}{|c}{\textbf{CVC-612}}\\ 
           \cline{2-11}  &  \multicolumn{1}{c|}{{mDice}}  & \multicolumn{1}{c|}{{$S_\alpha$}}  & \multicolumn{1}{c|}{{mDice}} & \multicolumn{1}{c|}{{$S_\alpha$}} & \multicolumn{1}{c|}{mDice}  &  \multicolumn{1}{c|}{{$S_\alpha$}}  & \multicolumn{1}{c|}{{mDice}}  & \multicolumn{1}{c|}{{$S_\alpha$}} & \multicolumn{1}{c|}{{mDice}} & \textbf{$S_\alpha$}  \\ \hline\hline
{2 NS-blocks}        & 0.944        & \multicolumn{1}{c|}{0.950}          & \multicolumn{1}{c|}{0.909}          & \multicolumn{1}{c|}{0.921}            &\multicolumn{1}{c|}{0.836}          & \multicolumn{1}{c|}{0.885}          & \multicolumn{1}{c|}{0.816}            & \multicolumn{1}{c|}{0.857}               &  \multicolumn{1}{c|}{0.899}               & 0.932                            \\ \hline
{3 NS-blocks (ours)}       &  \textbf{0.959}        & \multicolumn{1}{c|}{\textbf{0.962}}          & \multicolumn{1}{c|}{\textbf{0.925}}          & \multicolumn{1}{c|}{\textbf{0.938}}            & \multicolumn{1}{c|}{\textbf{0.841}}          & \multicolumn{1}{c|}{\textbf{0.899}}          & \multicolumn{1}{c|}{\textbf{0.827}}            & \multicolumn{1}{c|}{\textbf{0.871}}               &\multicolumn{1}{c|}{\textbf{0.911}}               & \textbf{0.942}                            \\ \hline
 {4 NS-blocks}        &  0.951        & \multicolumn{1}{c|}{0.957}          & \multicolumn{1}{c|}{0.916}          & \multicolumn{1}{c|}{0.929}            &\multicolumn{1}{c|}{0.832}          & \multicolumn{1}{c|}{0.881}          & \multicolumn{1}{c|}{0.820}            & \multicolumn{1}{c|}{0.864}               & \multicolumn{1}{c|}{0.906}               & 0.937                            \\ \hline
\end{tabular}
\end{table*}

\begin{table*}[t!h!]
\centering
\caption{Paired \textit{t}-test on different datasets. p-values (cutoff at $\geq 0.05$) demonstrates significant difference between SSTFB and other methods.\label{tab:ablation_pvalues}}
\begin{tabular}{l|c|c|c|c|c|c|c|c|c|c}
\hline
\textbf{comparison}  & \multicolumn{2}{c}{\textbf{Easy(seen)}} & \multicolumn{2}{|c}{\textbf{Hard(seen)}}& \multicolumn{2}{|c}{\textbf{Easy(unseen)}}& \multicolumn{2}{|c}{\textbf{Hard(unseen)}}& \multicolumn{2}{|c}{\textbf{CVC-612}}\\ 
           \cline{2-11}  &  \multicolumn{2}{c|}{{p-value}}  &  \multicolumn{2}{c|}{{p-value}} &  \multicolumn{2}{c|}{p-value}   & \multicolumn{2}{c|}{{p-value}}  &  \multicolumn{2}{c}{{p-value}}  \\ \hline\hline
{SSTFB \& PraNet}      & \multicolumn{2}{c|}{9.71E-14}          & \multicolumn{2}{c|}{8.03E-09}          & \multicolumn{2}{c|}{0.036}            &\multicolumn{2}{c|}{0.0195}          & \multicolumn{2}{c}{0.0327}                            \\ \hline
{SSTFB \& Polyp-PVT}       & \multicolumn{2}{c|}{7.59E-19}          & \multicolumn{2}{c|}{2.17E-10}          & \multicolumn{2}{c|}{0.0148}            &\multicolumn{2}{c|}{0.0096}          & \multicolumn{2}{c}{0.0219}                            \\ \hline
 {SSTFB \& PNS+}        & \multicolumn{2}{c|}{3.91E-56}          & \multicolumn{2}{c|}{6.46E-11}          & \multicolumn{2}{c|}{0.0077}            &\multicolumn{2}{c|}{0.00042}          & \multicolumn{2}{c}{0.0092}                            \\ \hline
\end{tabular}
\end{table*}

\subsection{Evaluation metrics}
Two conventional metrics, mean Dice Similarity Coefficient ($ mDice =\frac{2TP}{2TP+FP+FN}$) and mean Intersection over Union ($ mIoU =\frac{TP}{TP+FP+FN}$) to measure the similarity and overlap between predictions and ground truth are used. Here, TP is true positives, FP is false positives, and FN refers to the false negatives. 

In addition, we have also included the S-measure, which is the weighted ($\alpha$) metric between the object-aware structural similarity measure ($S_o$) and region-aware structural similarity measure ($S_r$)~\cite{fan2017structure} ($S_\alpha = \alpha\times S_o(P, G)+(1-\alpha)\times S_r(P, G)$). Here, $P$ is  prediction, $G$ is ground true, and $\alpha$ is set to 0.5. 

A weighted F-measure which corrects for ``Equal-importance flaws'' in Dice is also used~\cite{margolin2014evaluate, achanta2009frequency} ($F^w_\beta=(1+\beta^{2})\times precision^{w} \times recall^{w}/(\beta^{2} \times (precision^{w} + recall^{w}))$. Here, $precison = \frac{TP}{TP+FP}$, $recall = \frac{TP}{TP+FN}$, and $\beta^{2}=0.3$. 

Another metric referred to as an enhanced-alignment measure which is suitable for assessing heterogeneous location and shape of polyps in colonoscopy~\cite{ji2022video} is also used ($E_\phi = (1/(W \times H))\sum_{x=0}^W \sum_{y=0}^H \phi (P(x,y), G(x,y))$, $W$: width of $G$, $H$: height of $G$, $\phi$: enhanced alignment matrix)~\cite{fan2018enhanced}.
\subsection{Implementation details}
We train our SSTFB network using the PyTorch framework on a server deployed with NVIDIA Quadro RTX 6000 graphics cards. All input images were resized to $256\times 448$ pixels, and batch size was set to 24. The first frame of a video clip was chosen as an anchor and randomly sampled five consecutive frames as neighbouring frames ($\Delta=5$). Throughout the model training process, we set the initial learning rate and weight decay of the Adam optimiser to 3$e^{-4}$ and 1$e^{-4}$, respectively. The number of stacked normalised self-attention blocks is set as $R = 3$, the kernel size $k = 3$, and the number of attention groups to $N = 4$ with dilation rate of $\{3,4,3,4\}$ for the first two NS-blocks, and dilation rate of $\{1,2,1,2\}$ for the third NS-block. For the SSL task, we use the patches from the entire sequence of images fed to the Res2Net-50 model (pre-trained with ImageNet weights).\\



%
%
\section{Results}
\subsection{SOTA method comparisons}

In this section, we assess the performance of our method on two seen test (held-out) datasets (SUN-SEG-Easy(seen) \& SUN-SEG-Hard(seen)) and comparison with other SOTA methods.\\
\subsubsection{Quantitative results}

Quantitative results of SOTA methods and our proposed SSTFB are summarised in Table~\ref{tab:test-seen}. It can be observed from Table~\ref{tab:test-seen} that the performance of the method developed for still polyp image segmentation (PraNet, Polyp-PVT, SANet and TGANet) is lower than the video polyp image segmentation method (PNSNet, PNS+ and SSTFB(ours)) on both SUN-SEG-Easy and SUN-SEG-Hard datasets. Specifically, our proposed SSTFB reaches 0.911 on mDice, 0.856 on mIoU, $S_\alpha$ on 0.935, $E_\phi$ on 0.968 and $F^w_\beta$ on 0.871 achieving performance improvements of 9.9\%, 11.9\%, 7.8\%, 7.4\% and 12.8\% respectively compared with the polyp segmentation benchmark work (PraNet). SSTFB also achieves 7.8\% and 10.4\% increase of mDice and mIoU on SUN-SEG-Easy(seen) when contrasted with the recent SOTA method (TGANet). Simultaneously, our method still has a mDice improvement of about 3\% for the most recent SOTA polyp video segmentation model PNS+. On the SUN-SEG-Hard(seen) dataset, SSTFB presents mDice of 0.881 and mIoU of 0.813, which shows an over 8\% performance improvement compared with PraNet and 2.6\% improvement compared with PNS+.\\
Table~\ref{tab:fps} shows SSTFB can run at extremely high inference speeds (126fps). Our proposed SSTFB significantly outperforms all other SOTA models on all metrics with real-time performance. This shows that our model has a stronger learning ability and can effectively learn semantically meaningful features for polyp segmentation. \\

\subsection{Ablation Study}
We have conducted an extensive ablation study on five different datasets (SUN-SEG-Easy(seen), SUN-SEG-Hard(seen), SUN-SEG-Easy(unseen), SUN-SEG-Hard(unseen), and CVC-612) to evaluate the effectiveness of our proposed method. First, we ablate the impact of SSL block and feature branching. Moreover, we will conduct an extended ablation experiment to evaluate the performance of our approach further under different NS-Blocks settings. \\

\subsubsection{Effectiveness of SSL block.}
The external SSL block that uses SSL pre-training and the built-in SSL block in our end-to-end model are tested separately. From Table~\ref{tab:ablation_datasets}, we can observe that the end-to-end (e2e) model with built-in SSL block achieves 0.7\%, 0.5\%, 1.3\%, 0.9\% and 1.8\% of mDice improvement on five datasets. Meanwhile, the e2e model is also better than the external model (about 1\% improvement) in $S_\alpha$. This performance improvement shows that the joint training of the entire model through the end-to-end SSL block training can help the model to learn to capture semantically meaningful polyp features.\\

\subsubsection{Effectiveness of feature branching.}

In Table~\ref{tab:ablation_datasets}, the last row in the table represents the method's performance after using feature branching (SSL+e2e+b). Compared with other methods that are without feature branching, reported results have a significant performance improvement. Specifically, removal of feature branching shows a decrease in mice by 1.6\%, 2.3\%, 1.2\%, 1.2\% and 0.8\% and reductions in $S_\alpha$ by 1.2\%, 0.9\%, 2.1\%, 0.5\% and 1.1\% for SEG-Easy(seen), SUN-SEG-Hard(seen), SEG-Easy(unseen), SUN-SEG-Hard(unseen) and CVC-612. The results show that synthesising aggregated features by branching global and local features and input into soft attention blocks separately improves performance.\\

\subsubsection{Choice of NS-block.}
The choice of the number of NS-blocks with feature branching and fusion was also tested with our experiments demonstrating the current configuration to be the most optimal (See Table~\ref{tab:ablation_ns}). As shown in Table~\ref{tab:ablation_ns}, our method with three NS-blocks setting demonstrates superior performance, which has approximately a 1.5\% increase in both mDice and $S_\alpha$ on all datasets. The performance of model with four NS-blocks is slightly higher than with the two NS-blocks setting.\\

\subsection{Qualitative results}
\begin{figure}[t!h!]
    \centering
    \includegraphics[width=0.55\textwidth]{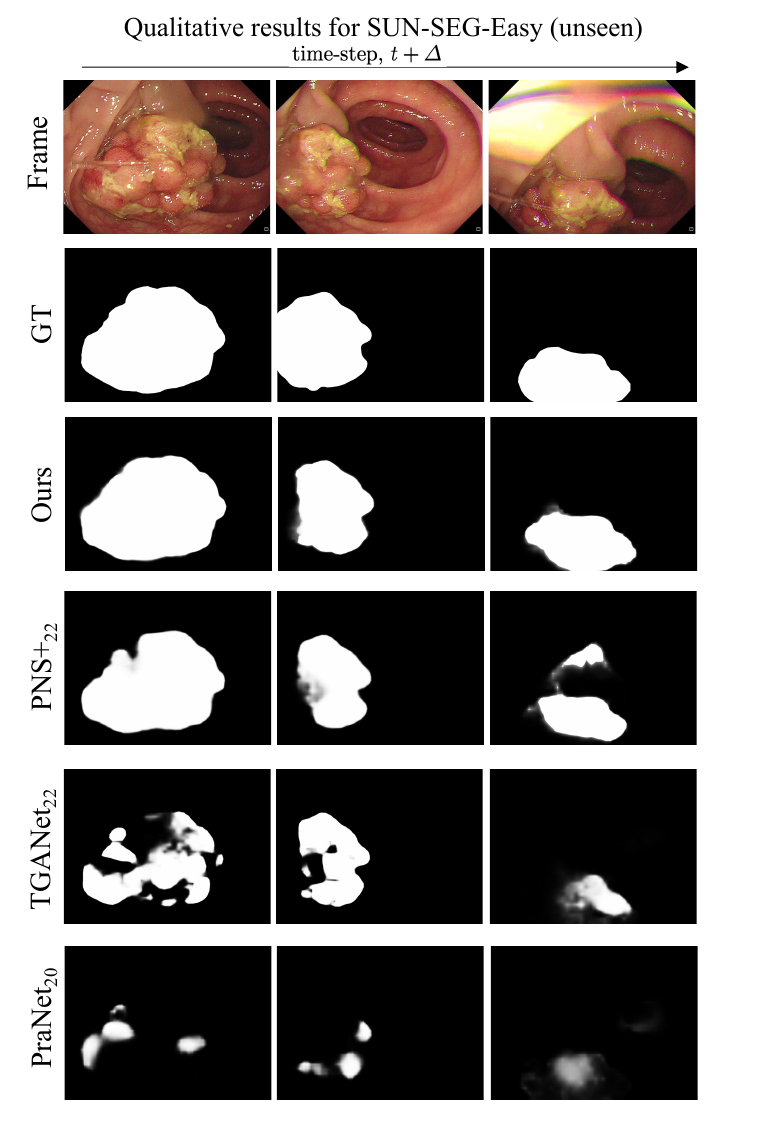}
    \caption{\textbf{Qualitative results:} Easy samples from SUN-SEG-Easy (unseen). }\label{fig:qualitative_result_easy}
\end{figure}

\begin{figure*}[t!h!]
    \centering
    \includegraphics[width=0.85\textwidth]{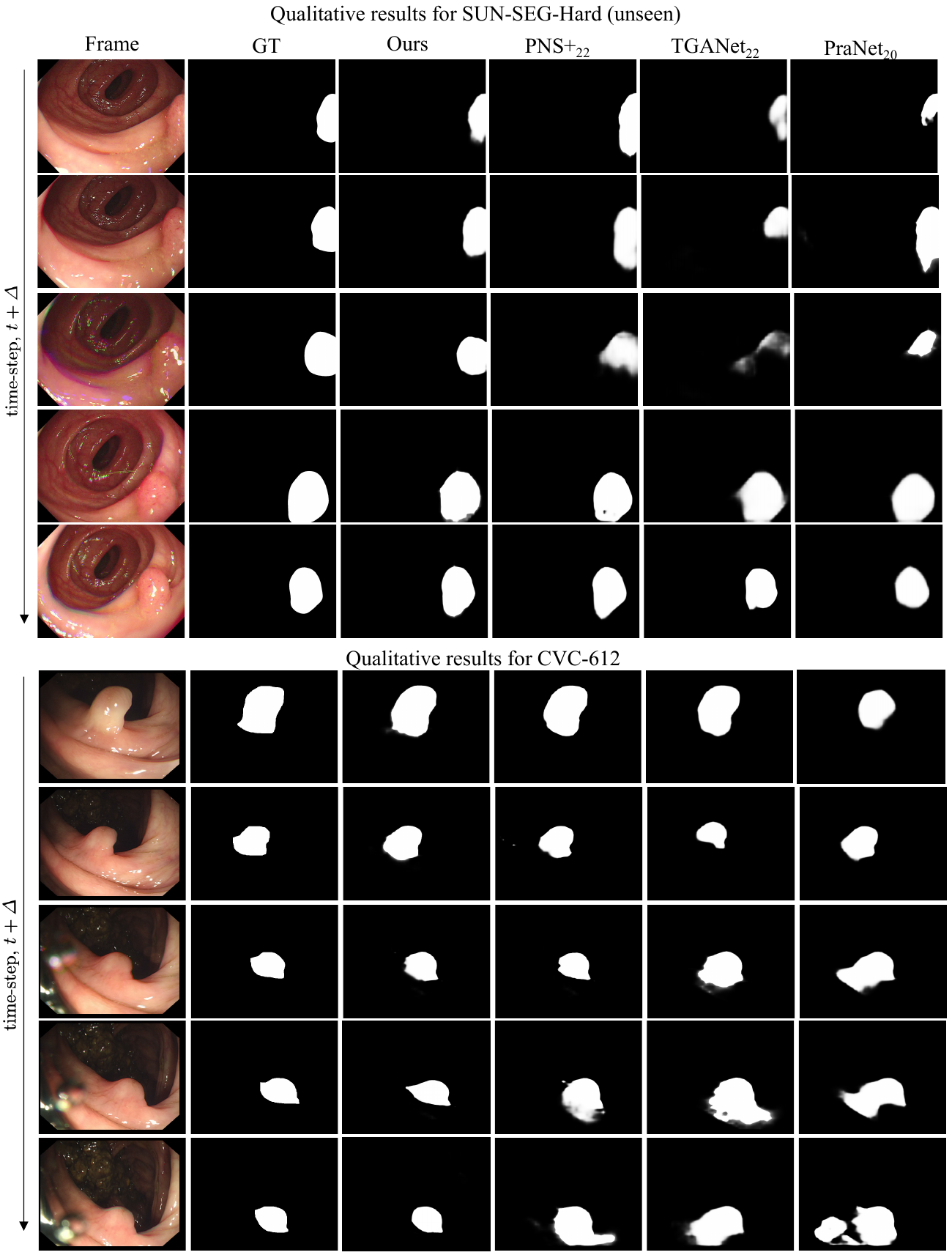}
    \caption{\textbf{Qualitative results:} Hard samples from SUN-SEG-Hard (unseen, top rows), and samples from unseen data centre CVC-612 (bottom rows). }\label{fig:qualitative_result_new}
\end{figure*}

Fig.~\ref{fig:qualitative_result_easy} and Fig.~\ref{fig:qualitative_result_new} demonstrate the effectiveness of our approach and competing networks for the video polyp segmentation on SUN-SEG and CVC-612 datasets. It can be observed that methods designed based on the video polyp segmentation task (SSTFB (Ours) and PNS+) show more obvious performances compared with methods applied to still polyp segmentation (PraNet and TGANet). Specifically, our proposed method demonstrates superior performance with sharper boundaries when dealing with multi-scale polyp segmentation and outperforms competing networks on various challenging video polyp segmentation tasks. For example, in the first to third columns of Fig.~\ref{fig:qualitative_result_easy}, competing methods vaguely locate polyps to a large extent, while SSTFB can accurately locate polyps. In the SUN-SEG-Hard part of Fig.~\ref{fig:qualitative_result_new}, most competing methods cannot accurately distinguish polyps from similar environments, which results in blurred edges of predicted polyps. Our method can cope with background interference and produce accurate edge information. Bubbles on the lens will also affect the judgment of the method to a certain extent. In the last row of the CVC-612 part of Fig.~\ref{fig:qualitative_result_new}, the competing method was affected by the bubbles and incorrectly judged the location of the polyp. Nonetheless, our method can still ignore the effect of bubbles and provide accurate segmentation results. Overall, the proposed SSTFB demonstrates a competitive advantage over other state-of-the-art networks in video polyp segmentation tasks.\\
\begin{figure}[t!h!]
    \centering
    \includegraphics[width=0.45\textwidth]{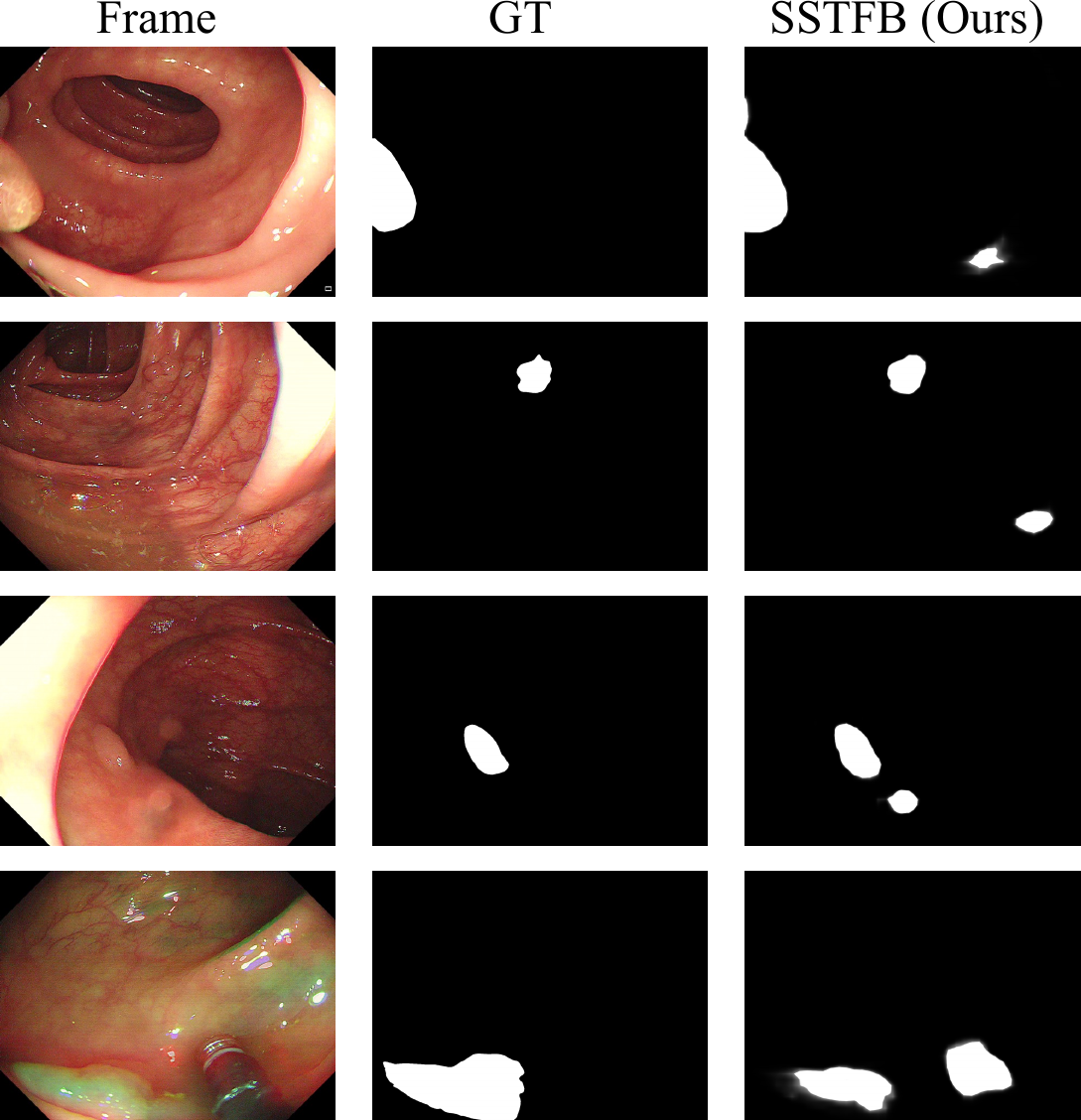}
    \caption{Limitations of our proposed approach (SSTFB), including bubbles, instruments, and imaging artefacts such as specularity and pixel saturation.}\label{fig:badcases}
\end{figure}
\section{Discussion}
Due to the similar environment of the internal structure of the intestine, frequent movement of the endoscope, and lighting issues, there are inevitable problems with similar disease areas to the intestinal wall in endoscopy videos, artefacts, saturation, and motion camera blur. Some images with lower scores are presented in Fig.~\ref{fig:badcases}. For example, in the first and third rows, the method is affected by flares and pixel saturation and incorrectly determines the polyp area. In addition, air bubbles (second row) and surgical instruments (fourth row) also cause false positive segmentation results. Additionally, the third row is affected by various colour artefacts (chromatic aberration) due to sudden motion changes or reflection. In these cases, a depressed region with such chromatic aberration is determined falsely as polyps. Mitigating data bias due to such variability as the causal effect in video data is important and can be tackled by video-based polyp segmentation only to some extent (Fig.~\ref{fig:qualitative_result_easy} and Fig.~\ref{fig:qualitative_result_new}), but it still outputs some false positives (Fig.~\ref{fig:badcases}). Compared to other methodologies developed based on static polyp images, approaches incorporating normalized self-attention blocks and a global-to-local learning strategy can more accurately segment polyp positions and contour information by leveraging both long-term and short-term spatial-temporal cues. Furthermore, integrating embedded SSL and sub-branching mechanisms can further learn meaningful representations and enhance the feature-separation capability of the normalized self-attention blocks. SSTFB has demonstrated superior segmentation results within both SUN-SEG-Easy(Fig.~\ref{fig:qualitative_result_easy}), SUN-SEG-Hard and CVC-612(Fig.~\ref{fig:qualitative_result_new}) datasets.

Furthermore, p-values from paired \textit{t}-tests represent significant differences between SSTFB and other approaches. It can be observed from Table~\ref{tab:ablation_pvalues} that our approach provided p-value $\leq 0.05$ (significantly different) compared with PraNet, Polyp-PVT and PNS+ in all five different datasets. Our network ablation on utilising an optimal number of NS blocks (Fig.~\ref{tab:ablation_ns}) also demonstrates that the chosen architecture is best suited for video polyp segmentation in contrast to other works previously published. 

Similarly, end-to-end learning of the video polyp segmentation, jointly fine-tuned with learned feature representation, SSL, and feature branching step, thereby segregating high-level and low-level features and their components, made the network robust to seen, unseen and out-of-distribution datasets (Table~\ref{tab:ablation_datasets}). It can be observed that performing self-supervised learning (pretext task for representation learning) does improve the result, but using the proposed combination in an end-to-end fashion favours learning. This also lowers the bottleneck of computational time requirement during training and provides more useful feature learning. As evident from our experiments, the proposed SSTFB has real-time performance (126FPS, Table~\ref{tab:fps}) with accuracies higher than all SOTA methods for polyp segmentation in the literature on SUN-SEG easy and hard samples. The improvement on seen (Table~\ref{tab:test-seen}) and unseen (Table~\ref{tab:test-unseeen}) samples ranges from nearly 10\% (PraNet) up to nearly 3\% (PNS+) on mDice metric. 
\section{Conclusion}
We proposed a novel end-to-end framework exploiting spatial-temporal information,  representation learning using self-supervision, and feature branching with normalised self-attention blocks, namely SSTFB. We provided detailed ablation of various network configurations that suggest that the proposed network outperforms existing SOTA methods without hurting performance time. Compared to many methods in SOTA that are trained and validated on single-frame endoscopy datasets, we demonstrated that our approach outperformed these methods by a larger margin and hence can be used for precise and real-time polyp segmentation in colonoscopy videos that are usually of natural occurrences in clinical settings. The proposed SSL block integration and the normalised self-attention block based on feature branching presented in this paper can be flexibly used in existing CNN-based segmentation models. Our extensive experiments showed the superiority of our method over all existing SOTA methods in all evaluation metrics, generalisability assessment, and visual quantification.  
\bibliographystyle{IEEEtran}
\vspace{-2mm}
\bibliography{refs} 
\end{document}